\documentclass[11pt,a4paper]{article}
\usepackage[]{naaclhlt2018}
\usepackage{times}
\usepackage{latexsym}
\usepackage{booktabs}

\usepackage{url}

\usepackage[]{todonotes}

\usepackage{soul}
\newcommand{\tabref}[1]{Table~\ref{#1}}
\newcommand{\figref}[1]{Figure~\ref{#1}}

\newcommand{\numdata}{ten\xspace}
\usepackage{multirow}
\usepackage{amsmath}

\usepackage{colortbl}
\usepackage{amssymb}
\usepackage{pifont}

\newcommand{\infersent}{\texttt{InferSent}\xspace}

\usepackage{linguex}

\usepackage{subfig}
\usepackage{caption}
\usepackage{multirow}

\usepackage{dcolumn}
\newcolumntype{H}{>{\setbox0=\hbox\bgroup}c<{\egroup}@{}}

\usepackage{xspace}
\usepackage{color}

\usepackage{enumitem}
\newenvironment{datalist}
{\begin{itemize}[leftmargin=0mm]}
{\end{itemize}}

\aclfinalcopy 
\def\aclpaperid{163} 

\newcommand\capsize{\capsizeinner}

\title{Hypothesis Only Baselines in Natural Language Inference}
\author{Adam Poliak$^{1}$ \hspace{.1cm} Jason Naradowsky$^{1}$ \hspace{.1cm} Aparajita Haldar$^{1,2}$ \\ \textbf{Rachel Rudinger$^{1}$ \hspace{.1cm} Benjamin Van Durme$^{1}$} \\
$^{1}$Johns Hopkins University 
$^{2}$BITS Pilani, Goa Campus, India
\\
{\normalsize \tt \{azpoliak,vandurme\}@cs.jhu.edu} 
{\normalsize \tt \{narad,ahaldar1,rudinger\}@jhu.edu}
}
\date{}

\begin{document}
\maketitle
\begin{abstract}
We propose a \emph{hypothesis only} baseline for diagnosing Natural Language Inference (NLI).  Especially when an NLI dataset assumes inference is occurring based purely on the relationship between a context and a hypothesis, it follows that assessing entailment relations while ignoring the provided context is a degenerate solution.  Yet, through experiments on \numdata distinct NLI datasets, we find that this approach, which we refer to as a hypothesis-only model, is able to significantly outperform a majority-class baseline across a number of NLI datasets.  Our analysis suggests that statistical irregularities may allow a model to perform NLI in some datasets beyond what should be achievable without access to the context.
\end{abstract}

\section{Introduction}
Though datasets for the task of Natural Language Inference (NLI) may vary in just about every aspect (size, construction, genre, label classes), they generally share a common structure: each instance consists of two fragments of natural language text (a \textit{context}, also known as a \textit{premise}, and a \textit{hypothesis}), and a label indicating the entailment relation between the two fragments (e.g., \textsc{entailment}, \textsc{neutral}, \textsc{contradiction}).
Computationally, the task of NLI is to predict an entailment relation label (output) given a premise-hypothesis pair (input), i.e., to determine whether the truth of the hypothesis follows from the truth of the premise \cite{dagan2006pascal,dagan2013recognizing}.

\iftrue
\begin{figure}[t]
\centering
\subfloat[]{\includegraphics[width=.66\linewidth,trim=0 0 0 0,clip,scale=1.5]{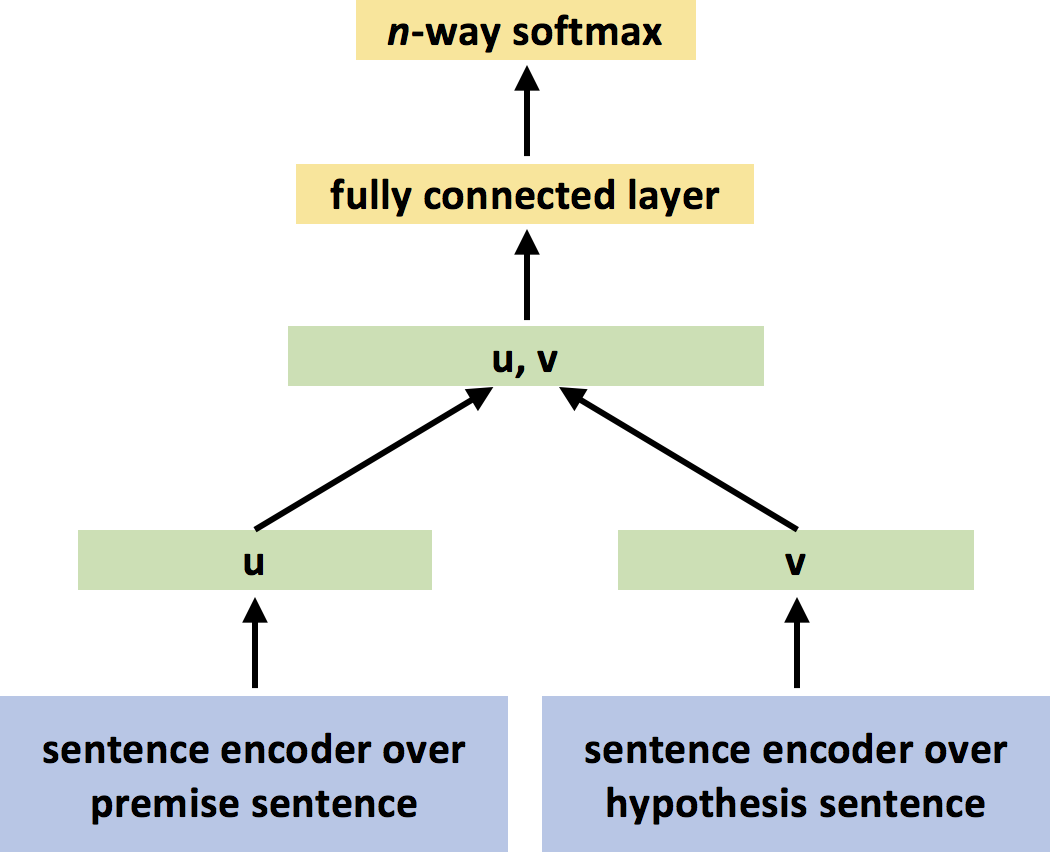} \label{fig:both}}
\subfloat[]{\includegraphics[width=.32\linewidth,trim=0 0 0 0,clip,scale=1.5]{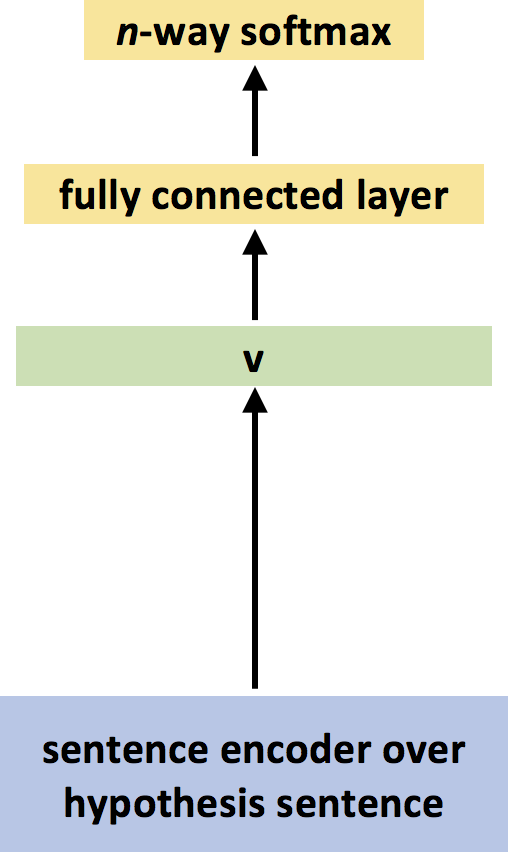} \label{fig:hyp}}
\label{fig:ff1}
\caption{\capsize (\ref{fig:both}) shows a typical NLI model that encodes the premise and hypothesis sentences
into a vector space to classify the sentence pair. (\ref{fig:hyp}) shows our hypothesis-only
baseline method that ignores the premise and only encodes the hypothesis sentence.}
\end{figure}
\fi

When these NLI datasets are constructed to facilitate the training and evaluation of  natural language understanding (NLU) systems
\cite{nangia2017repeval}, it is tempting to claim that systems achieving high accuracy on such datasets have successfully ``understood'' natural language or at least a logical relationship between a premise and hypothesis.
While this paper does not attempt to prescribe the sufficient conditions of such a claim, we argue for an obvious \textit{necessary}, or at least \textit{desired} condition: \textit{that interesting natural language inference should depend on both premise and hypothesis.} In other words, a baseline system with access only to 
hypotheses~(\figref{fig:hyp}) can be said to perform NLI only in the sense that it is understanding language based on prior background knowledge.
If this background knowledge is about the world, this may be justifiable as an aspect of natural language understanding, if not in keeping with the spirit of NLI.
But if the ``background knowledge" consists of learned statistical irregularities in the data, this may not be ideal.
Here we explore the question: do NLI datasets contain 
statistical irregularities that allow hypothesis-only models to outperform the datasets specific prior?

We present the results of a hypothesis-only baseline across \numdata NLI-style datasets and advocate for its inclusion in future dataset reports.
We find that this baseline 
can perform above the majority-class prior across most of the \numdata examined datasets.  We examine whether: (1) hypotheses contain statistical irregularities within each entailment class that are ``giveaways'' to a well-trained hypothesis-only model,
(2) the way in which an NLI dataset is constructed is related to how prone it is to this particular weakness, 
and
(3) the majority baselines might not be as indicative of ``the difficulty of the task''~\cite{snli:emnlp2015}
as previously thought.

We are not the first to consider the inherent
difficulty of NLI datasets.
For example, \newcite{maccartney2009natural}
used a simple bag-of-words model
to evaluate early iterations of Recognizing Textual Entailment (RTE) challenge sets.\footnote{\newcite{maccartney2009natural}, Ch. 2.2: \emph{``the RTE1 test suite is the hardest, while the RTE2 test suite is roughly 4\%
easier, and the RTE3 test suite is roughly 9\% easier.''}} Concerns have been raised previously about the hypotheses in the Stanford Natural Language Inference (SNLI) dataset specifically,  such as by \newcite{social-bias-in-elicited-natural-language-inferences}
and in unpublished work.\footnote{A course project constituting independent discovery of our observations on SNLI: \url{https://leonidk.com/pdfs/cs224u.pdf}}
Here, we survey of large number of existing NLI datasets under the lens of a hypothesis-only model.\footnote{
Our code and data can be found at
\url{https://github.com/azpoliak/hypothesis-only-NLI}.}
Concurrently, \newcite{1804.08117} and \newcite{Gururangan:2018} similarly trained an NLI classifier with access limited to hypotheses and discovered similar results on three of the ten datasets that we study.

\section{Motivation}
Our approach is inspired by recent studies that show how biases in an NLU dataset allow models to perform well on the task without understanding the meaning of the text.  In the Story Cloze task~\cite{mostafazadeh-EtAl:2016:N16-1, mostafazadeh-EtAl:2017:LSDSem}, a model is presented with a short four-sentence narrative and asked to complete it by choosing one of two suggested concluding sentences.  While the task is presented as a new common-sense reasoning framework, \newcite{schwartz2017story} achieved state-of-the-art performance by ignoring the narrative and training a linear classifier with features related to the writing style of the two potential endings, rather than their content.  It has also been shown that features focusing on sentence length, sentiment, and negation are sufficient for achieving high accuracy on this dataset~\cite{schwartz2017effect,Cai2017PayAT,bugert2017lsdsem}.

NLI is often viewed as an integral part of NLU. \newcite{condoravdi2003entailment} argue that it is a necessary metric for evaluating an NLU system, since it forces a model to perform many distinct types of reasoning.  \newcite{goldberg2017neural} suggests that ``solving [NLI]
perfectly entails human level understanding of language'', and \newcite{nangia2017repeval} argue that ``in order for a system to perform well at natural language inference,
it needs to handle nearly the full complexity
of natural language understanding.''  However, if biases in NLI datasets, especially those that do not reflect commonsense knowledge, allow models to achieve high levels of performance without needing to reason about hypotheses based on corresponding contexts, our current datasets may fall short of these goals.

\section{Methodology}

We modify \newcite{conneau-EtAl:2017:EMNLP2017}'s \infersent method to train a neural model to classify just the hypotheses.  We choose \infersent because it performed competitively with the best-scoring systems on the Stanford Natural Language Inference (SNLI) dataset~\cite{snli:emnlp2015}, while being representative of the types of neural architectures commonly used for NLI tasks. \infersent uses a BiLSTM encoder, and constructs a sentence representation by max-pooling over its hidden states. This sentence representation of a hypothesis is
used as input to a MLP classifier to predict the NLI tag.

We preprocess 
each recast dataset
using the NLTK 
tokenizer~\cite{loper2002nltk}. 
Following \newcite{conneau-EtAl:2017:EMNLP2017}, we map the resulting tokens to 300-dimensional GloVe vectors~\cite{pennington2014glove}
trained on 840 billion tokens from the Common Crawl, using the GloVe OOV vector for unknown words.
We optimize via SGD, with an initial learning rate of $0.1$, and decay rate of $0.99$.
We allow at most $20$ epochs of training with optional early stopping according to the following policy:
when the accuracy on the development set decreases, we divide the learning rate by $5$ and stop training when learning rate is $<$ $10^{-5}$.

\section{Datasets}

We collect \numdata NLI datasets and categorize them into three distinct groups
based on the methods by which they were constructed.  \tabref{tab:nli-datasets} summarizes the different NLI datasets that our
investigation considers.

\begin{table*}
\centering
\begin{tabular}[t!]{clrcl}
\toprule
\textbf{Creation Protocol} &  \textbf{Dataset} & \textbf{Size} & \textbf{Classes} & \multicolumn{1}{c}{\textbf{Example Hypothesis}} \\
\midrule
\multirow{3}{*}{Recast} & DPR   & 3.4K  & 2 &  \textit{\small{People raise dogs because dogs are afraid of thieves}}
\\ 
&  SPR  & 150K & 2 & \textit{\small{The judge was 
aware of the dismissing }}\\
& FN+   & 150K   & 2 & \textit{\small{the irish are actually principling to come home}}\\
\midrule
\multirow{5}{*}{Judged} & ADD-1 & 5K &  2 &  \textit{\small{A small child staring at a young horse and a pony}} \\
& SCITAIL & 25K  & 2 & \textit{\small{Humans typically have 23 pairs of chromosomes}}\\
& SICK & 10K &   3 & \textit{\small{Pasta is being put into a dish by a woman}}\\
& MPE & 10K &  3 &  \textit{\small{A man smoking a cigarette }}\\
&  JOCI & 30K   & 3 &  \textit{\small{The flooring is a horizontal surface}}\\
\midrule
\multirow{2}{*}{Elicited}  & SNLI  & 550K  &  3 &  \textit{\small{An animal is jumping to catch an object}} \\
&  MNLI  & 425K  & 3 &  \textit{\small{Kyoto has a kabuki troupe and so does Osaka}}\\
\bottomrule
\end{tabular}
\caption{\capsize
Basic statistics about the NLI datasets we consider.
`Size' refers to the total number of labeled premise-hypothesis 
pairs in each dataset (for datasets with $>100K$ examples,  numbers are rounded down to the nearest $25K$).
The `Creation Protocol' column indicates how
the dataset was created. The `Class' column reports the number of class labels/tags. The last column shows an example hypothesis from each dataset.
}
\label{tab:nli-datasets}
\end{table*}

\subsection{Human Elicited}
In cases where humans were given a context and asked to generate a corresponding hypothesis and label,
we consider these datasets to be \textbf{elicited}.   Although we consider only two such datasets, they are the largest datasets included in our study and are currently popular amongst researchers.  The elicited NLI datasets we look at are:

\begin{datalist}
\item[] \textbf{Stanford Natural Language Inference (SNLI)}
To create SNLI,
\newcite{snli:emnlp2015} showed crowdsourced workers a premise sentence (sourced from Flickr image captions), and asked them to generate a corresponding hypothesis sentence for each of the three labels (\textsc{entailment}, \textsc{neutral}, \textsc{contradiction}).  SNLI is known to contain stereotypical biases based on gender, race, and ethnic 
stereotypes~\cite{social-bias-in-elicited-natural-language-inferences}.
Furthermore, \newcite{TACL1082} commented that this ``elicitation protocols can
lead to biased responses unlikely to contain a wide
range of possible common-sense inferences.''

\item[] \textbf{Multi-NLI}
Multi-NLI is a recent expansion of SNLI 
aimed to add greater diversity to the existing dataset~\cite{williams2017broad}.  
Premises in Multi-NLI can originate from fictional stories, personal letters, telephone speech, and a 9/11 report.

\end{datalist}

\subsection{Human Judged}
Alternatively,
if hypotheses and premises were automatically paired but \emph{labeled} by a human, we consider the dataset to be \textbf{judged}.  Our human-judged data sets are:

\begin{datalist}
\item[] \textbf{Sentences Involving
Compositional Knowledge (SICK)}
To evaluate how well compositional distributional semantic models
handle ``challenging phenomena'', \newcite{MARELLI14.363.L14-1314} introduced SICK, which used rules to expand or normalize existing premises to create more difficult examples.
Workers were asked to label the relatedness of these resulting pairs, and these labels were then converted into the same three-way label space as SNLI and Multi-NLI.

\item[] \textbf{Add-one RTE}
This mixed-genre dataset tests whether NLI systems can understand adjective-noun compounds~\cite{P16-1204}.
Premise sentences were extracted from
Annotated Gigaword~\cite{napoles-gormley-vandurme:2012:AKBC-WEKEX},
image captions~\cite{young2014image}, the Internet Argument Corpus~\cite{walker2012stance},
and fictional stories from the GutenTag dataset~\cite{mac2015finding}.
To create hypotheses, adjectives were removed or inserted before nouns in a premise, and crowd-sourced workers were asked to provide reliable labels (\textsc{entailed}, \textsc{not-entailed}).

\item[] \textbf{SciTail}
Recently released, SciTail is an NLI dataset created from $4$th grade science questions and multiple-choice answers~\cite{scitail}.
Hypotheses are assertions converted from
question-answer pairs found in SciQ~\cite{welbl2017crowdsourcing}.   
Hypotheses are automatically paired with premise sentences from domain specific texts~\cite{clark2016combining}, and labeled (\textsc{entailment}, \textsc{neutral}) by crowdsourced workers.  Notably, the construction method allows for the same sentence to appear as a hypothesis for more than one premise.

\item[] \textbf{Multiple Premise Entailment (MPE)}
Unlike the other datasets we consider, the premises in MPE~\cite{lai-bisk-hockenmaier:2017:I17-1} are not single sentences, but four different captions that describe the same image in the FLICKR30K dataset~\cite{plummer2015flickr30k}.
Hypotheses were generated by simplifying either a fifth caption that describes the
same image or a caption corresponding to a different image, and given the standard 3-way tags.
Each hypothesis has at most a 50\% overlap with the
words in its corresponding premise.
Since the hypotheses are still just one sentence, our hypothesis-only baseline 
can easily be applied to MPE.

\item[] \textbf{Johns Hopkins Ordinal Common-Sense Inference (JOCI)}
JOCI 
labels context-hypothesis instances
on an ordinal scale from \textit{impossible} ($1$) to \textit{very likely} ($5$)~\cite{TACL1082}. In JOCI,
context (premise) sentences were taken from existing NLU datasets: SNLI, ROC Stories~\cite{mostafazadeh-EtAl:2016:N16-1}, and COPA~\cite{roemmele2011choice}. 
Hypotheses were created automatically by 
systems trained to generate entailed facts from a premise.\footnote{
We only consider the hypotheses generated by either a seq2seq model or from external world knowledge.}
Crowd-sourced
workers 
labeled the likelihood of the hypothesis following from the premise on an \emph{ordinal scale}.  We convert these into a $3$-way NLI tags where 1 maps to \textsc{contradiction}, 2-4 maps to \textsc{neutral}, and 5 maps to \textsc{entailment}.
Converting the annotations into a $3$-way classification problem 
allows us to limit the range of the number of NLI label classes 
in our investigation.

\end{datalist}

\subsection{Automatically Recast}
If an NLI dataset was automatically generated from existing datasets for other NLP tasks, and sentence pairs were constructed and labeled 
with minimal human intervention, 
we refer to such a dataset as \textbf{recast}. We use the  recast datasets from \newcite{white-EtAl:2017:I17-1}:

\begin{datalist}
\item[] \textbf{Semantic Proto-Roles (SPR)}
Inspired by \newcite{dowty1991thematic}'s thematic role theory, \newcite{TACL674}
introduced the Semantic Proto-Role (SPR) labeling task, which can be viewed as decomposing semantic roles into finer-grained properties, such as whether a predicate's argument was likely \emph{aware} of the given predicated situation. 2-way labeled NLI sentence pairs were generated from SPR annotations by creating 
general templates.

\definecolor{light-gray}{gray}{0.97}

\begin{table*}
\centering
\small
\begin{tabular}[t!]{ c ccrr  ccrr  cc }
\toprule
& \multicolumn{4}{c}{\textbf{DEV}} & \multicolumn{4}{c}{\textbf{TEST}} & \\
Dataset & Hyp-Only & MAJ & $ \centering|\Delta|$ & $\centering\Delta$\% & Hyp-Only & MAJ & $\centering|\Delta|$ &\centering$\Delta$\%  & Baseline & SOTA \\
\cmidrule(l){2-5}   \cmidrule(l){6-9}
\multicolumn{11}{c}{\cellcolor{light-gray} Recast} \\
\midrule
\textit{DPR}   & 50.21  & 50.21 & 0.00 & 0.00 & 49.95 & 49.95 & 0.00 & 0.00 & 49.5 & 49.5 \\
SPR & 86.21 & 65.27 & \textcolor{blue}{+20.94} & \textcolor{blue}{+32.08} & 86.57 & 65.44 & \textcolor{blue}{+21.13} & \textcolor{blue}{+32.29} & 80.6 & 80.6 \\
FN+ & 62.43 & 56.79 &\textcolor{blue}{+5.64} &\textcolor{blue}{+9.31} & 61.11 & 57.48 & \textcolor{blue}{+3.63} & \textcolor{blue}{+6.32} & 80.5 & 80.5 \\
\midrule
\multicolumn{11}{c}{\cellcolor{light-gray} Human Judged} \\
\textit{ADD-1} & 75.10 & 75.10 & 0.00 & 0.00 & 85.27 & 85.27 & 0.00  & 0.00 & 92.2 & 92.2 \\
SciTail & 66.56 & 50.38 & \textcolor{blue}{+16.18} & \textcolor{blue}{+32.12} & 66.56 & 60.04 &  \textcolor{blue}{+6.52} & \textcolor{blue}{+10.86} & 70.6 & 77.3 \\
\textit{SICK} & 56.76 & 56.76 & 0.00 & 0.00 & 56.87 & 56.87 & 0.00 & 0.00 & 56.87 & 84.6 \\
\textit{MPE}  & 40.20 & 40.20 & 0.00 & 0.00 & 42.40 & 42.40 & 0.00 & 0.00 & 41.7 & 56.3 \\
JOCI & 61.64 & 57.74 & \textcolor{blue}{+3.90} & \textcolor{blue}{+6.75} & 62.61 & 57.26 & \textcolor{blue}{+5.35} & \textcolor{blue}{+9.34} & -- & -- \\
\midrule
\multicolumn{11}{c}{\cellcolor{light-gray} Human Elicited} \\
SNLI & 69.17 & 33.82 & \textcolor{blue}{+35.35} & \textcolor{blue}{+104.52} & 69.00 & 34.28 & \textcolor{blue}{+34.72} & \textcolor{blue}{+101.28} & 78.2 & 89.3 \\
MNLI-1 & 55.52 & 35.45 & \textcolor{blue}{+20.07} & \textcolor{blue}{+56.61} & -- & 35.6 & -- -- & & 72.3 & 80.60\\
MNLI-2 & 55.18 & 35.22 & \textcolor{blue}{+19.96}  & \textcolor{blue}{+56.67}  & -- &  36.5 & -- & -- & 72.1 & 83.21 \\
\bottomrule
\end{tabular}
\caption{\capsize NLI accuracies 
on each dataset.
Columns `Hyp-Only' and `MAJ' indicates the accuracy of the hypothesis-only model and the majority baseline.
$|\Delta|$ and $\Delta$\% indicate the absolute difference in percentage points and the percentage increase between the Hyp-Only and MAJ. Blue numbers indicate that the hypothesis-model outperforms MAJ.
In the right-most section, `Baseline' indicates the original baseline on the test when the dataset was released and `SOTA' indicates current state-of-the-art results.
MNLI-1 is the matched version and MNLI-2 is the mismatched for MNLI.
The names of datasets are italicized if containing $\leq10K$ labeled examples.
}
\label{tab:nli-res-all}
\end{table*}

\item[] \textbf{Definite Pronoun Resolution (DPR)}
The DPR dataset
targets an NLI model's ability
to perform anaphora resolution~\cite{rahman-ng:2012:EMNLP-CoNLL}.
In the original dataset, sentences contain two entities and one pronoun, and the task is to link the pronoun to its referent. In the recast version, the premises are the original sentences and
the hypotheses are the same sentences with the pronoun replaced with its correct (\textsc{entailed}) and incorrect (\textsc{not-entailed}) referent.
For example, 
\textit{People raise dogs because they are obedient} and \textit{People raise dogs because dogs are obedient}
is such a context-hypothesis pair.
We note that this mechanism would appear to maximally benefit a
hypothesis-only approach, as the hypothesis semantically subsumes the
context.

\item[] \textbf{FrameNet Plus (FN+)}
Using paraphrases from PPDB~\cite{ganitkevitch2013ppdb}, \newcite{rastogi2014augmenting}
automatically replaced words with their paraphrases. Subsequently,
\newcite{pavlick-EtAl:2015:ACL-IJCNLP2} asked crowd-source workers to judge
how well a sentence with 
a paraphrase preserved the original sentence's meanings.
In this NLI dataset that targets a model's ability to perform paraphrastic inference,
premise sentences are the original sentences, the hypotheses are the edited versions, and the 
crowd-source judgments are converted to 2-way NLI-labels. For not-entailed examples, \newcite{white-EtAl:2017:I17-1} replaced a single token in a context sentence with a word that crowd-source workers labeled as not being a paraphrase of the token in the given context. In turn, we might suppose that positive entailments~\ref{fnplus-correct} are keeping in the spirit of NLI, but not-entailed examples might not because there are adequacy~\ref{fnplus-adequacy} and fluency~\ref{fnplus-fluency} issues.\footnote{In these examples,~\ref{fnplus-context} is the corresponding context.}

\ex.
\a. That is the way the system works
\label{fnplus-context}
\b. That is the way the framework works
\label{fnplus-correct}
\c. That is the road the system works 
\label{fnplus-adequacy}
\d. That is the way the system creations
\label{fnplus-fluency}

\end{datalist}

\section{Results}
Our goal is to determine whether a hypothesis-only model outperforms the majority baseline and investigate what may cause significant gains.
In such cases a hypothesis-only model should be used as a stronger baseline instead of the majority class baseline. 
For all experiments except for JOCI, we use each NLI dataset's standard train, dev, and test
splits.\footnote{JOCI was not released with such splits so we randomly split the dataset into such a partition with 80:10:10 ratios.}
\tabref{tab:nli-res-all} compares the hypothesis-only model's accuracy with the majority baseline on each dataset's dev and test set.\footnote{We only report results on the Multi-NLI development set since the test labels are only accessible on Kaggle.}

\paragraph{Criticism of the Majority Baseline}
Across six of the \numdata datasets, our hypothesis-only model \textit{\textbf{significantly outperforms}} the majority-baseline, even outperforming the best reported results on one dataset, recast SPR.  This indicates that there exists a significant degree of exploitable signal
that may help NLI models perform well on their corresponding test set without considering NLI contexts.  From \tabref{tab:nli-res-all}, it is unclear whether the construction method is responsible for these improvements. 
The largest relative gains are on human-elicited models where the hypothesis-only model more than doubles
the majority baseline.

However, there are no obvious unifying trends across these datasets:  Among the judged and recast datasets, where humans do not generate the NLI hypothesis, we observe lower performance margins between majority and hypothesis-only models compared to the elicited data sets.  However, the baseline performances of these models are noticeably larger than on SNLI and Multi-NLI.  The drop between SNLI and Multi-NLI suggests that by including multiple genres, an NLI dataset may contain less biases. However, adding additional genres might not be enough to mitigate biases as the hypothesis-only model still drastically outperforms the majority-baseline. Therefore, we believe that models tested on SNLI and Multi-NLI should include a baseline version of the model that only accesses hypotheses. 

We do not observe general trends across the datasets based on their construction methodology. On three of the five human judged datasets, the hypothesis-only model defaults to labeling each instance with the majority class tag.  We find the same behavior in one recast dataset (DPR).  However, across both these categories we find smaller relative improvements than on SNLI and Multi-NLI.
These results suggest the existence of exploitable signal in the datasets that is unrelated to NLI contexts. 
Our focus now shifts to identifying precisely what these signals might be and understanding why they may appear in NLI hypotheses.

\section{Statistical Irregularities}
We are interested in determining what characteristics in the datasets may be responsible for the hypothesis-only model often outperforming the majority baseline. Here, we investigate 
the importance of specific words,  grammaticality, and lexical semantics.

\begin{figure*}[t]
\centering
\subfloat[][SNLI]{\includegraphics[width=.5\linewidth,trim=0 0 0 0,clip,scale=.5]{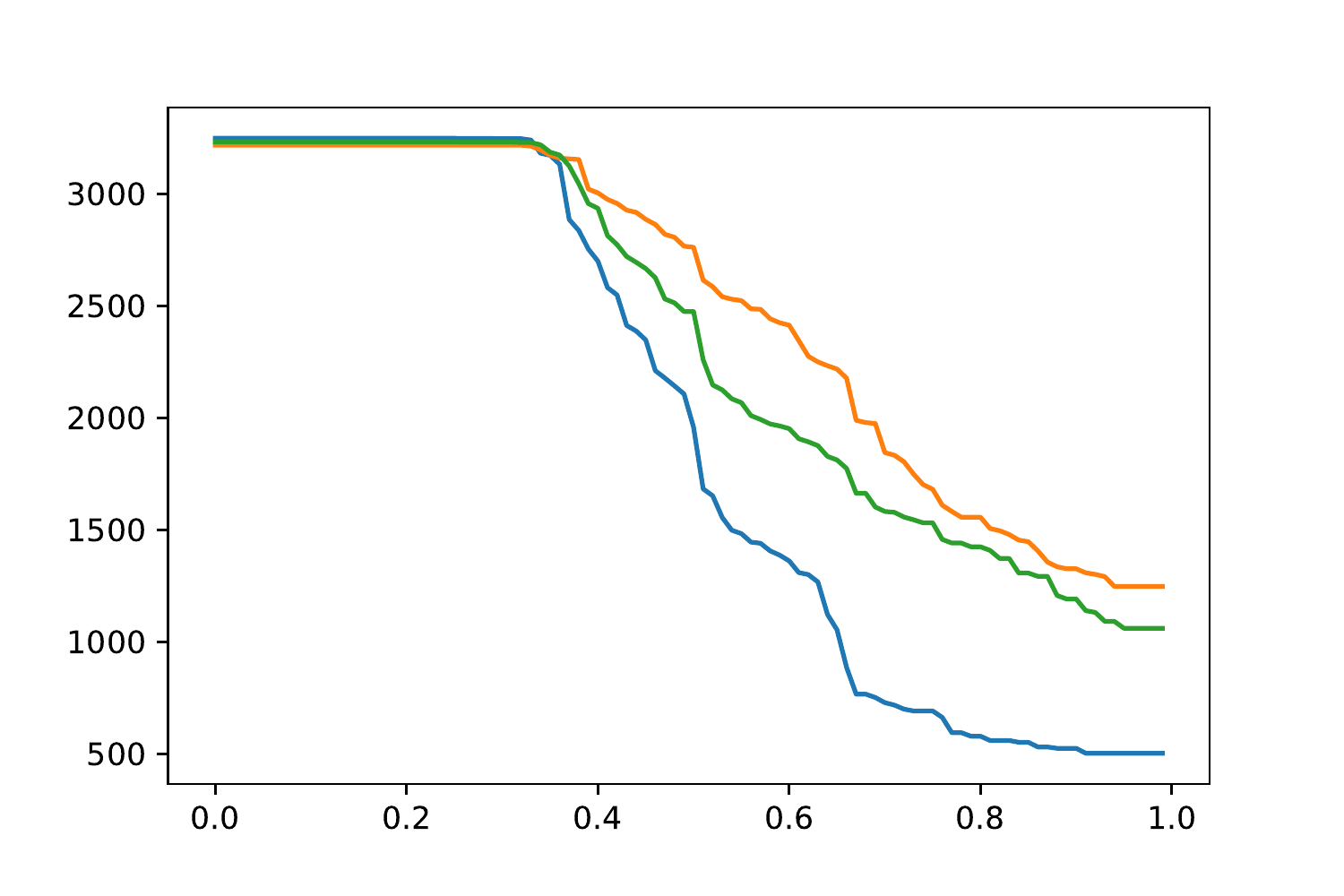}
\label{fig:snli}}
\subfloat[][DPR]
{\includegraphics[width=.5\linewidth,trim=0 0 -10 0,clip,scale=.5]{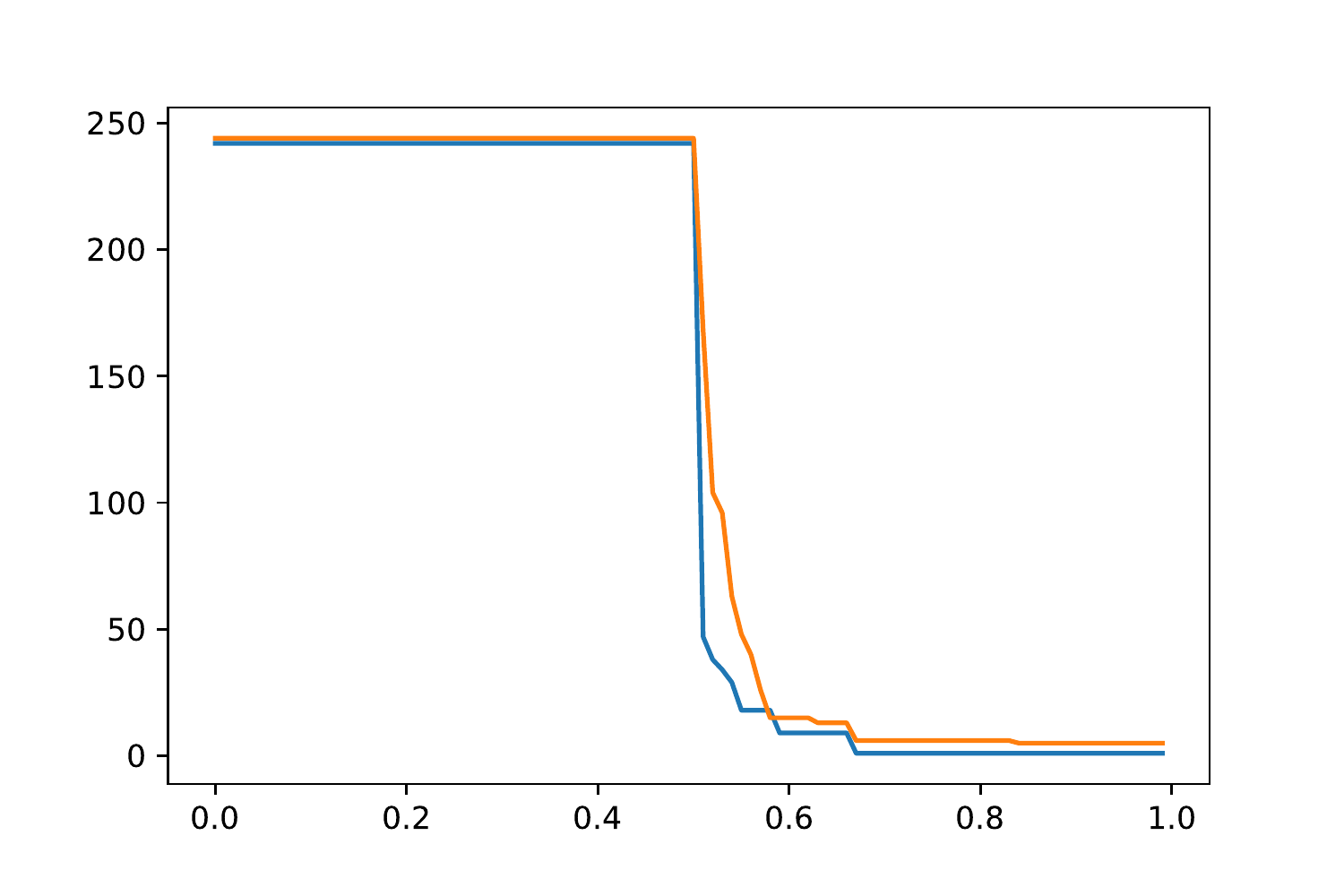}
\label{fig:scitail}}
\caption{\capsize Plots showing the number of sentences per each label (Y-axis) that contain at least one word $w$ such that $p(l|w) >= x$ for at least one label $l$.  Colors indicate different labels.  Intuitively, for a sliding definition of what value of $p(l|w)$ might constitute a ``give-away'' the Y-axis shows the proportion of sentences that can be trivially answered for each class.}
\label{fig:word-sent-importance}
\end{figure*}

\subsection{Can Labels be Inferred from Single Words?}
Since words in hypotheses have a distribution over the class of labels, we can determine the conditional probability of a label $l$ given the word $w$
by 
\begin{equation}
p(l|w) = \dfrac{count(w,l)}{count(w)}
\label{eq:p-of-l-given-w}
\end{equation}
If $p(l|w)$ is highly skewed across labels, there exists the potential for a predictive bias.  Consequently, such words may be ``give-aways'' that allow the hypothesis model to correctly predict an NLI label without considering the context.

If a single occurrence of a highly label-specific word would allow a sentence to be deterministically classified, how many sentences in a dataset are prone to being trivially labeled?  The plots in \figref{fig:word-sent-importance} answer this question for SNLI and DPR.  The $Y$-value where $X=1.0$ captures the number of such sentences.  Other values of $X<1.0$ can also have strong correlative effects, but a priori the relationship between the value of $X$ and the coverage of trivially answerable instances in the data is unclear.  We illustrate this relationship for varying values of $p(l|w)$.
When $X=0$, all words are considered highly-correlated with a specific class label, and thus the entire data set would be treated as trivially answerable.

In DPR, which has two class labels, because
the uncertainty of a label is highest when $p(l|w) = 0.5$, the sharp drop as $X$ deviates from this value indicates a weaker effect, where there are proportionally fewer sentences which contain highly label-specific words with respect to SNLI.  As SNLI uses 3-way classification we see a gradual decline from 0.33. 

\subsection{What are ``Give-away'' Words?}

Now that we 
analyzed the extent to which 
highly label-correlated words may exist across sentences in a given label, we would like to understand what these words are and why they exist.

~\figref{fig:word-lists} reports some of the words with the highest $p(l|w)$ for SNLI,  a human elicited dataset, and MPE, a human judged dataset, on which our hypothesis model performed identically to the majority baseline.
Because many of the most discriminative words are low frequency, we report only words which occur at least five times.
We rank the words according to their overall frequency, since this statistic is perhaps more indicative of a word $w$'s effect on overall performance
compared to $p(l|w)$ alone.

The score $p(l|w)$ of the words shown for SNLI deviate strongly, regardless of the label.  In contrast, in MPE, 
scores are much closer to 
a uniform distribution of $p(l|w)$ across labels.  Intuitively, the stronger the word's deviation, 
the stronger the potential for it to be a ``give-away'' word.  
A high word frequency 
indicates a greater potential of the word to affect the overall accuracy on NLI.
\begin{figure*}[t]
\centering
\subfloat[][entailment]{
\begin{tabular}[h]{c c c}
\toprule
\\
\toprule
\textbf{Word} & \textbf{Score} & \textbf{Freq} \\
\midrule
instrument & 0.90 & 20 \\
touching & 0.83 & 12 \\
least & 0.90 & 10 \\
Humans & 0.88 & 8 \\
transportation & 0.86 & 7 \\
speaking & 0.86 & 7 \\
screen & 0.86 & 7 \\
arts & 0.86 & 7 \\
activity & 0.86 & 7 \\
opposing & 1.00 & 5 \\
\bottomrule
\end{tabular}
}
\subfloat[][neutral]{
\begin{tabular}[h]{c c c}
\toprule
\multicolumn{3}{c}{\textbf{SNLI}}\\
\toprule
\textbf{Word} & \textbf{Score} & \textbf{Freq} \\
\midrule
tall & 0.93 & 44 \\
competition & 0.88 & 24 \\
because & 0.83 & 23 \\
birthday & 0.85 & 20 \\
mom & 0.82 & 17 \\
win & 0.88 & 16 \\
got & 0.81 & 16 \\
trip & 0.93 & 15 \\
tries & 0.87 & 15 \\
owner & 0.87 & 15 \\
\bottomrule
\end{tabular}
}
\subfloat[][contradiction]{
\begin{tabular}[h]{c c c}
\toprule
\\
\toprule
\textbf{Word} & \textbf{Score} & \textbf{Freq} \\
\midrule
sleeping & 0.88 & 108 \\
driving & 0.81 & 53 \\
Nobody & 1.00 & 52 \\
alone & 0.90 & 50 \\
cat & 0.84 & 49 \\
asleep & 0.91 & 43 \\
no & 0.84 & 31 \\
empty & 0.93 & 28 \\
eats & 0.83 & 24 \\
sleeps & 0.95 & 20 \\
\bottomrule
\end{tabular}
}
\\
\subfloat[][entailment]{
\begin{tabular}[h]{c c c}
\toprule
\\
\toprule
\textbf{Word} & \textbf{Score} & \textbf{Freq} \\
\midrule
an & 0.57 & 21 \\
gathered & 0.58 & 12 \\
girl & 0.50 & 12 \\
trick & 0.55 & 11 \\
Dogs & 0.55 & 11 \\
watches & 0.60 & 10 \\
field & 0.60 & 10 \\
singing & 0.50 & 10 \\
outside & 0.67 & 9 \\
something & 0.62 & 8 \\
\bottomrule
\end{tabular}
}
\subfloat[][neutral]{
\begin{tabular}[h]{c c c}
\toprule
\multicolumn{3}{c}{\textbf{MPE}} \\
\toprule
\textbf{Word} & \textbf{Score} & \textbf{Freq} \\
\midrule
smiling & 0.56 & 16 \\
An & 0.60 & 10 \\
for & 0.56 & 9 \\
front & 0.75 & 8 \\
camera & 0.62 & 8 \\
waiting & 0.50 & 8 \\
posing & 0.50 & 8 \\
Kids & 0.57 & 7 \\
smile & 0.83 & 6 \\
wall & 0.50 & 6 \\
\bottomrule
\end{tabular}
}
\subfloat[][contradiction]{
\begin{tabular}[h]{c c c}
\toprule
\\
\toprule
\textbf{Word} & \textbf{Score} & \textbf{Freq} \\
\midrule
sitting & 0.51 & 88 \\
woman & 0.55 & 80 \\
men & 0.56 & 34 \\
Some & 0.62 & 26 \\
doing & 0.59 & 22 \\
Children & 0.50 & 22 \\
boy & 0.67 & 21 \\
having & 0.65 & 20 \\
sit & 0.60 & 15 \\
children & 0.53 & 15 \\
\bottomrule
\end{tabular}
}
\caption{\capsize Lists of the most highly-correlated words in each dataset for given labels, thresholded to the top 10 and ranked according to frequency.}
\label{fig:word-lists}
\end{figure*}

\noindent \paragraph{Qualitative Examples}
Turning our attention to the qualities of the words themselves, we can easily identify trends among the words used in contradictory hypotheses in SNLI.  In our top-10 list, for example, three words refer to the act of sleeping.  Upon inspecting corresponding context sentences, we find that many contexts, which are sourced from Flickr, naturally deal with activities.  This leads us to believe that as a common strategy, crowd-source workers often do not generate contradictory hypotheses that require fine-grained semantic reasoning, as a majority of such activities can be easily negated by removing an agent's agency, i.e. describing the agent as sleeping.
A second trend we notice is that universal negation constitutes four of the remaining seven terms in this list, and may also be used to similar effect.\footnote{These are ``Nobody'', ``alone'', ``no'', and ``empty''.} The human-elicited protocol does not guide, nor incentivize crowd-source workers to come up with less obvious examples.
If not properly controlled, elicited datasets may be prone to many label-specific terms.
The existence of label-specific terms in human-elicited NLI datasets does not invalidate the datasets nor is surprising. Studies in eliciting norming data are prone to repeated responses across subjects~\cite{mcrae2005semantic} (see discussion in \S 2 of \cite{TACL1082}).

\subsection{On the Role of Grammaticality}
Like MPE, FN+ contains few high frequency words with high $p(l|w)$.
However, unlike on MPE,
our hypothesis-only model outperforms the majority-only baseline.  If these gains 
do not arise from ``give-away'' words, then
what is the statistical irregularity responsible for this discriminative power?

Upon further inspection, we notice an interesting imbalance in how our model performs for each of the two classes. The hypothesis-only model performs similarly to the majority baseline for entailed examples, while improving by over 34\% those which are not entailed, as shown in~\tabref{tab:fn-plus-breakdown}.

\begin{table}
\centering
\small
\begin{tabular}[h!]{r ccHc}
\toprule
\textbf{label} & \textbf{Hyp-Only} & \textbf{MAJ} & \textbf{$|\Delta|$} & \textbf{$\Delta$\%} \\
entailed & 44.18 & 43.20 & \textcolor{blue}{+0.98}  & \textcolor{blue}{+2.27} \\
not-entailed & 76.31 & 56.79 & \textcolor{blue}{+19.52} & \textcolor{blue}{+34.37} \\
\bottomrule
\end{tabular}
\caption{\capsize Accuracies on FN+ for each class label.}
\label{tab:fn-plus-breakdown}
\end{table}

As shown by \newcite{white-EtAl:2017:I17-1} and noticed by \newcite{poliakNAACL18}, FN+ contains more grammatical errors than the other recast datasets.  We explore whether grammaticality could be the statistical irregularity exploited in this case.
We manually sample a total of $200$ FN+ sentences and categorize them based on their gold label and our model's prediction. Out of $50$ sentences that the model correctly labeled as \textsc{entailed}, 88\% of them were grammatical. On the other-hand, of the $50$ hypotheses incorrectly labeled as \textsc{entailed}, only $38$\% of them were grammatical. Similarly, when the model correctly labeled $50$ \textsc{not-entailed}
hypotheses, only $20\%$ were grammatical, and $68\%$ when labeled incorrectly. This 
suggests that a hypothesis-only model may be able to discover the correlation between grammaticality and NLI labels on this dataset.

\subsection{Lexical Semantics}

A survey of gains~(\tabref{tab:spr-breakdown-dev}) in the SPR dataset suggest a number of its property-driven hypotheses, such as \emph{X was sentient in [the event]}, can be accurately guessed based on  lexical semantics (background knowledge learned from training) of the argument.  For example, the hypothesis-only baseline correctly predicts the truth of hypotheses in the dev set such as: \emph{Experts were sentient ...} or \emph{Mr. Falls was sentient ...}, and the falsity of \emph{The campaign was sentient}, while failing on referring expressions like \emph{Some} or \emph{Each side}. 
A model exploiting regularities of the real world would seem to be a different category of dataset bias: while not strictly \emph{wrong} from the perspective of NLU, one should  be aware of what the hypothesis-only baseline is capable of,  to recognize those cases where access to the context is required and therefore more interesting under NLI.

\subsection{Open Questions}

There may  remain statistical irregularities, which we leave for future work to explore.  For example, are there correlation between sentence length and label class in these data sets?  Is there a particular construction method that minimizes the amount of ``give-away'' words present in the dataset?  And lastly, our study is another in a line of research which looks for irregularities at the word level~\cite{maccartney2008phrase, maccartney2009natural}.  Beyond bag-of-words, are there 
multi-word expressions or syntactic phenomena that might encode label biases?

\begin{table}
\centering
\small
\begin{tabular}[htpb!]{ c c c c }
\toprule
\small{\textbf{Proto-Role}}	& \small{\textbf{H-model}}	& \small{\textbf{MAJ}}	& \small{\textbf{$\Delta$\%}} \\
\midrule
\small{aware}	& 88.70	& 59.94 & \textcolor{blue}{+47.99} \\
\small{used in}	& 77.30	& 52.72 & \textcolor{blue}{+46.63} \\
\small{volitional}	& 87.45	& 64.96 & \textcolor{blue}{+34.62} \\
\small{physically existed}	& 87.97	& 65.38 & \textcolor{blue}{+34.56} \\
\small{caused}	& 82.11	& 63.08 & \textcolor{blue}{+30.18} \\
\small{sentient}	& 94.35	& 76.26 & \textcolor{blue}{+23.73} \\
\small{existed before}	& 80.23	& 65.90 & \textcolor{blue}{+21.75} \\
\small{changed}	& 72.18	& 64.85 & \textcolor{blue}{+11.29} \\
\small{chang. state}	& 71.76	& 64.85 & \textcolor{blue}{+10.65} \\
\small{existed after}	& 79.29	& 72.91 & \textcolor{blue}{+8.75} \\
\small{existed during}	& 90.06	& 85.67 & \textcolor{blue}{+5.13} \\
\small{location}	& 93.83	& 91.21 & \textcolor{blue}{+2.87} \\
\small{physical contact}	& 89.33	& 86.92 & \textcolor{blue}{+2.77} \\
\small{chang. possession}	& 94.87	& 94.46 & \textcolor{blue}{+0.44} \\
\small{moved}	& 93.51	& 93.20 & \textcolor{blue}{+0.34} \\
\small{stationary during}	& 96.44	& 96.34 & \textcolor{blue}{+0.11} \\
\bottomrule
\end{tabular}
\caption{\capsize NLI accuracies on the SPR development data;  each property appears in $956$ hypotheses.}
\label{tab:spr-breakdown-dev}
\end{table}

\section{Related Work}

\noindent
\paragraph{Non-semantic information to help NLI}
In NLI datasets, non-semantic linguistic features have been used to improve NLI models. \newcite{vanderwende2006syntax} and \newcite{Blake:2007:RSS:1654536.1654557} demonstrate how sentence structure alone can provide a high signal for NLI. Instead of using external sources of knowledge, which was a common trend at the time, \newcite{Blake:2007:RSS:1654536.1654557} improved results on RTE by combining syntactic features. More recently,
\newcite{bar2015knowledge} introduce an inference formalism based on syntactic-parse trees.

\noindent
\paragraph{World Knowledge and NLI}
As mentioned earlier, hypothesis-only models that perform without exploiting statistical irregularities may be performing NLI only in the sense that it is understanding language based on prior background knowledge. Here, we take the approach that \textit{interesting} NLI should depend on both premise and hypotheses. Prior work in NLI reflect this approach. For example, \newcite{glickman2005probabilistic-lexical-coocurrence} argue that ``the notion of textual entailment is relevant only'' for hypothesis that are not world facts, e.g. ``Paris is the capital of
France.'' \newcite{glickman2005probabilistic-lexical-te,glickman2005probabilistic},
introduce a probabilistic framework for NLI
where the premise entails a hypothesis if, and only if, the probability of the hypothesis being true increases as a result of the premise.

\noindent
\paragraph{NLI's resurgence}
Starting in the mid-2000's, multiple community-wide shared tasks focused on NLI,
then commonly referred to as RTE, i.e, recognizing textual entailment. 
Starting with \newcite{dagan2006pascal}, there have been eight iterations of the PASCAL RTE challenge
with the most recent being \newcite{dzikovska-EtAl:2013:SemEval-2013}.\footnote{Technically \newcite{bentivogli2011seventh}
was the last challenge under PASCAL's aegis but \newcite{dzikovska-EtAl:2013:SemEval-2013} was branded as the $8$th RTE challenge.}
NLI datasets were relatively small, ranging from 
\textit{thousands} 
to \textit{tens of thousands} of labeled sentence pairs.
In turn, NLI models often used alignment-based techniques~\cite{maccartney2008phrase} or
manually engineered features~\cite{androutsopoulos2010survey}.
\newcite{snli:emnlp2015} sparked a renewed interested in NLI, particularly among deep-learning researchers.
By developing and releasing a large NLI dataset containing over $550K$ examples, \newcite{snli:emnlp2015} enabled the community to successfully apply
deep learning models to the NLI problem.

\section{Conclusion}
We introduced a stronger baseline for \numdata NLI datasets.  Our baseline reduces the task from labeling the relationship between two sentences to classifying a single hypothesis sentence.  Our experiments demonstrated that in six of the \numdata datasets, always predicting the majority-class label is not a strong baseline, as it is significantly outperformed by the hypothesis-only model. 
Our analysis suggests that statistical irregularities, including word choice and grammaticality, may reduce the difficulty of the task on popular NLI datasets by not fully testing how well a model can determine whether the truth of a hypothesis follows from the truth of a corresponding premise.

We hope our findings will encourage the development of new NLI datasets which exhibit less exploitable irregularities, and that encourage the development of richer models of inference. As a baseline, new NLI models should be compared against a corresponding version that only accesses hypotheses.
In future work, we plan to apply a similar hypothesis-only baseline to multi-modal tasks that attempt to challenge a system to understand and classify the relationship between two inputs, e.g. Visual QA \cite{antol2015vqa}.

\section*{Acknowledgements}
This work was supported by Johns Hopkins University,
the Human Language Technology Center
of Excellence (HLTCOE),
DARPA LORELEI, and the NSF Graduate Research Fellowships Program (GRFP). We would also like to thank
three anonymous reviewers for their feedback.
The views and conclusions contained in this publication
are those of the authors and should not be
interpreted as representing official policies or endorsements
of DARPA or the U.S. Government.

\bibliography{references}
\bibliographystyle{acl_natbib}

\end{document}